\documentclass[sigconf]{acmart}

\usepackage{amsmath,amsfonts,bm}

\def\eqref#1{equation~\ref{#1}}

\def\1{\bm{1}}

\def\mD{{\bm{D}}}

\def\mI{{\bm{I}}}

\def\mP{{\bm{P}}}

\def\mS{{\bm{S}}}

\def\mU{{\bm{U}}}
\def\mV{{\bm{V}}}

\def\mX{{\bm{X}}}
\def\mY{{\bm{Y}}}
\def\mZ{{\bm{Z}}}

\DeclareMathAlphabet{\mathsfit}{\encodingdefault}{\sfdefault}{m}{sl}
\SetMathAlphabet{\mathsfit}{bold}{\encodingdefault}{\sfdefault}{bx}{n}

\usepackage[font=small,skip=0pt]{caption}
\usepackage{graphicx,adjustbox}
\usepackage[linesnumbered,ruled,vlined]{algorithm2e}
\SetKwComment{Comment}{/* }{ */}
\SetAlFnt{\small}
\SetKwInput{kwInput}{Input}
\SetKwInput{kwOutput}{Output}
\def\code#1{\texttt{#1}}
\AtBeginDocument{%
  \providecommand\BibTeX{{%
    \normalfont B\kern-0.5em{\scshape i\kern-0.25em b}\kern-0.8em\TeX}}}

\setcopyright{acmlicensed}
\copyrightyear{2024}
\acmYear{2024}
\acmDOI{XXXXXXX.XXXXXXX}

\acmConference[KDD 2024]{30th ACM SIGKDD Conference on Knowledge Discovery and Data Mining}{Aug 25 - 29, 2024}{Barcelona, Spain}
\acmISBN{978-1-4503-XXXX-X/18/06}

\usepackage{enumitem}
\begin{document}

\title{ShapeFormer: Shapelet Transformer for Multivariate Time Series Classification}

\author{Xuan-May Le}
\email{xuanmay.le@student.unimelb.edu.au}
\affiliation{%
  \institution{The University of Melbourne}
  \city{Melbourne}
  \state{Victoria}
  \country{Australia}
  \postcode{3010}
}

\author{Ling Luo}
\email{ling.luo@unimelb.edu.au}
\affiliation{%
  \institution{The University of Melbourne}
  \city{Melbourne}
  \state{Victoria}
  \country{Australia}
  \postcode{3010}
}

\author{Uwe Aickelin}
\email{uwe.aickelin@unimelb.edu.au}
\affiliation{%
  \institution{The University of Melbourne}
  \city{Melbourne}
  \state{Victoria}
  \country{Australia}
  \postcode{3010}
}

\author{Minh-Tuan Tran}
\email{tuan.tran7@monash.edu}
\affiliation{%
  \institution{Monash University}
  \city{Melbourne}
  \state{Victoria}
  \country{Australia}
  \postcode{3800}
}

\renewcommand{\shortauthors}{Le, Xuan-May et al.}

\begin{abstract}

Multivariate time series classification (MTSC) has attracted significant research attention due to its diverse real-world applications. Recently, exploiting transformers for MTSC has achieved state-of-the-art performance. However, existing methods focus on generic features, providing a comprehensive understanding of data, but they ignore class-specific features crucial for learning the representative characteristics of each class. This leads to poor performance in the case of imbalanced datasets or datasets with similar overall patterns but differing in minor class-specific details. In this paper, we propose a novel Shapelet Transformer (ShapeFormer), which comprises class-specific and generic transformer modules to capture both of these features. In the class-specific module, we introduce the discovery method to extract the discriminative subsequences of each class (i.e. shapelets) from the training set. We then propose a Shapelet Filter to learn the difference features between these shapelets and the input time series. We found that the difference feature for each shapelet contains important class-specific features, as it shows a significant distinction between its class and others. In the generic module, convolution filters are used to extract generic features that contain information to distinguish among all classes. For each module, we employ the transformer encoder to capture the correlation between their features. As a result, the combination of two transformer modules allows our model to exploit the power of both types of features, thereby enhancing the classification performance. Our experiments on 30 UEA MTSC datasets demonstrate that ShapeFormer has achieved the highest accuracy ranking compared to state-of-the-art methods. The code is available at \url{https://github.com/xuanmay2701/shapeformer}.

\end{abstract}




 \maketitle

\section{Introduction}

\begin{figure}[t]
\begin{center}
\includegraphics[width=\linewidth]{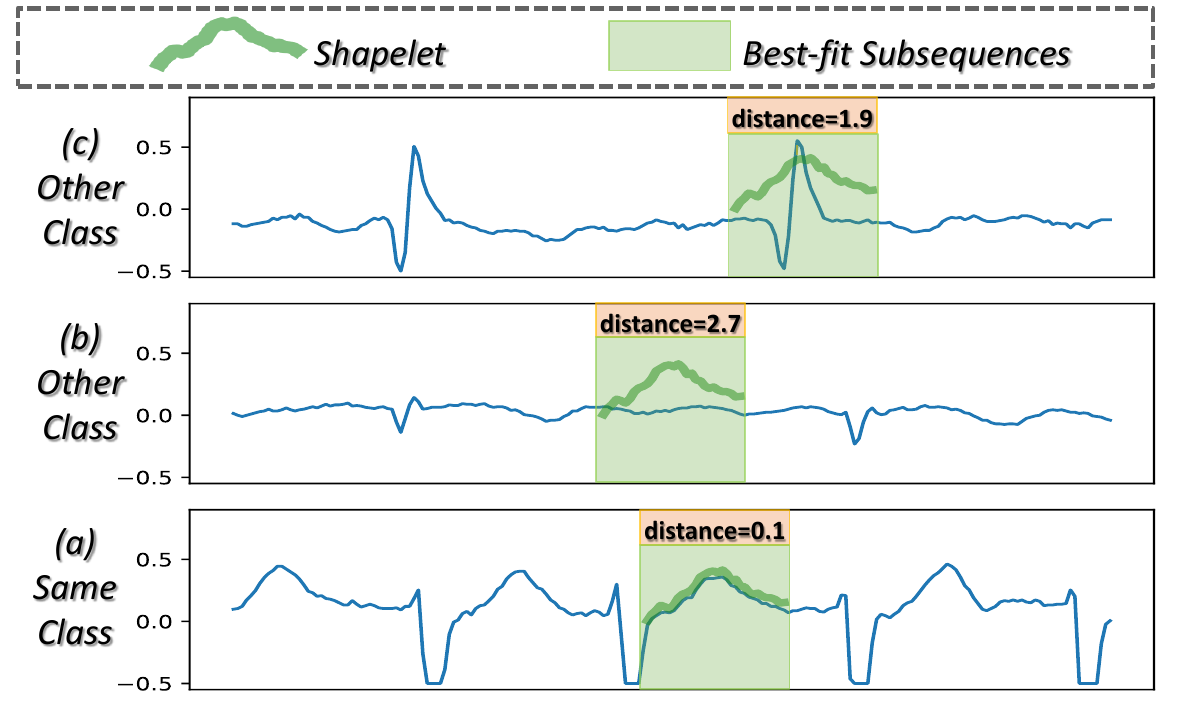}
\end{center}
\caption{The illustration depicts the shapelet in the Atrial Fibrillation dataset. The best-fit subsequence is the subsequence with the sortest distance to the shapelet in the time series. It is clear that the shapelet can discriminate between classes by utilising their distance to the best-fit subsequences.}
\label{fig:shapelets}
\end{figure}

A multivariate time series (MTS) is a collection of data points where each point is composed of multiple variables that have been observed or measured over time. This data structure is prevalent in various fields, such as economics \cite{economic}, weather prediction \cite{weather}, education \cite{education}, and healthcare \cite{healthcare}. Time series classification stands out as a fundamental and crucial aspect within the domain of time series analysis \cite{survey-tsc}. However, there are still many challenges in the research on MTS classification (MTSC) \cite{survey-tsc}, especially in capturing the correlations among variables. 

Over the past few decades, various approaches have been introduced to enhance the performance of MTSC \cite{tapnet, weasel, shapenet, when, RLPAM, svp-t}. Among these, shapelets, which are class-specific time series subsequences, have demonstrated their effectiveness in \cite{ppsn, shapenet, shapelet, fastshapelet-mtc}. This success comes from the fact that each shapelet contains class-specific information representative of its class. It is evident that the distance between the shapelet and the time series of its class is far smaller than the time series of other classes (see Figure \ref{fig:shapelets}). Hence, there has been an increased focus on harnessing the capabilities of shapelets in the field of MTSC.

In 2017, Vaswani et al. \cite{transformer} introduced the breakthrough Transformer architecture, initially designed for Natural Language Processing but later demonstrating success in Computer Vision tasks \cite{vision-transformer}. Following these successes, Transformer-based models have been effectively applied to MTSC. GTN \cite{gatedtrans} employs a two-tower multi-headed attention approach to extract distinctive information from input series, SVP-T \cite{svp-t} captures short- and long-term dependencies among subseries using clustering and employing them as inputs for the Transformer, and ConvTran \cite{ConvTran} integrates absolute and relative position encoding for improved position embedding in the Transformer model.

\begin{figure}[t]
\begin{center}
\includegraphics[width=0.95\linewidth]{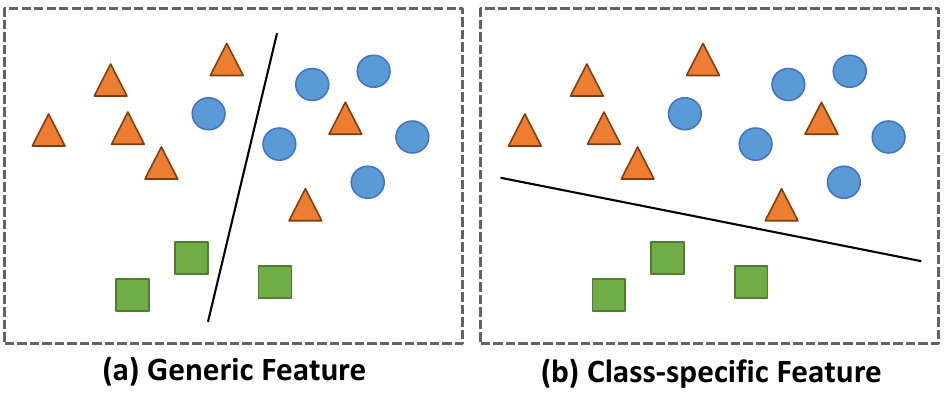}
\end{center}
\caption{The separating hyperplane using (a) the generic feature has a higher overall accuracy, while the hyperplane using (b) the class-specific feature is better in classifying a single class. }
\label{fig:gen-spe}
\end{figure}

Obviously, Transformers utilised in MTSC have demonstrated state-of-the-art (SOTA) performances \cite{TST, svp-t, ConvTran}. Existing methods only discover the generic features from {timestamps} \cite{TST, gatedtrans, ConvTran} or {common subsequences} \cite{svp-t} in time series as inputs for the Transformer model to capture the correlation among them. These features merely contain generic characteristics of time series, offering a broad understanding of the data. Nevertheless, they overlook the essential class-specific features necessary to allow the model to capture the representative characteristics of each class. As a result, the model exhibits poor performance in two cases: 1) the dataset has instances that are very similar in overall patterns, differing only in minor class-specific patterns, effective classification cannot be achieved using solely generic features; 2) the imbalanced dataset, where generic features only focus on classifying the majority classes and ignore those of minority. As can be seen in Figure \ref{fig:gen-spe}, the hyperplane created using the generic feature (Figure \ref{fig:gen-spe}a) attempts to classify the majority classes (orange triangles and blue circles) and ignores the minority (green squares), while the class-specific feature (Figure \ref{fig:gen-spe}b) tries to separate each class from the others.

To address the aforementioned problem, we propose a novel method called Shapelet Transformer (ShapeFormer), which comprises class-specific and generic transformer modules to capture both of these features. In the class-specific module, we initially introduce Offline Shapelet Discovery, inspired by \cite{ppsn}, to MTS. Based on this, we extract a small number of high-quality shapelets from the training set. Subsequently, we propose a Shapelet Filter that leverages the precomputed shapelets to discover the best-fit subsequences in the input time series. Following this, the Shapelet Filter learns the difference between the embedding of these shapelets and their most fitting subsequences derived from the input time series. As shown in Figure \ref{fig:shapelets}, the distance of shapelets to the time series in the same class is far smaller than the time series of other classes. Similar to the distance, our difference feature also highlights the substantial distinctions among classes. Additionally, rather than using the original shapelets extracted from the dataset, we propose considering these shapelets as the initialisation and then dynamically optimising shapelets during training to effectively represent the distinguishing information. In the generic module, we utilise convolution filters for the extraction of features over all classes. For each module, we employ the transformer encoder to capture the dependencies between their features. Through the integration of these two modules, our ShapeFormer excels in capturing not only class-specific features but also generic characteristics from time series data. This dual capability contributes to an enhancement in the overall performance of classification tasks.

Our contributions can be summarised as follows:
\begin{itemize}[noitemsep, nolistsep,leftmargin=*]
    \item We introduce ShapeFormer, which effectively captures both class-specific and generic discriminative features in time series.
    \item We propose the Offline Shapelet Discovery for MTS to effectively and efficiently extract shapelets from training set.
    \item We propose the Shapelet Filter, which learns the difference between shapelets and input time series, which contain important class-specific features. The shapelets are also dynamically optimised during training to effectively represent the class distinguishing information.
    \item We conduct experiments on all 30 UEA MTS datasets and demonstrate that ShapeFormer has achieved the highest accuracy ranking compared to SOTA methods.
\end{itemize}

To the best of our knowledge, our ShapeFormer is a pioneering transformer-based approach that leverages the power of shapelets for MTSC.

\section{Relative Works}
\subsection{Multivariate Time Series Classification}

We categorise the MTSC methods into two main categories: non-deep learning, and deep learning.

\noindent
\textbf{Non-deep learning methods.} They primarily utilise distance measures \cite{wknn, wldbaknn}, such as Euclidean Distance \cite{ed}, Dynamic Time Warping, and its diverse variants \cite{dtw1, dtw2}, to calculate the similarity between time series. Otherwise, they leverage special features, such as bag of patterns \cite{bag-of-patterns}, Symbolic Aggregate approXimation \cite{csax}, bag of SFA symbols \cite{boss}, and convolution kernel features \cite{minirocket, multirocket} for classification. \cite{survey} gives a comprehensive survey of the conventional methods mentioned. 

\noindent
\textbf{Deep learning methods}. Various neural network methods were proposed for MTSC \cite{deepreview}. Specifically, the LSTM-FCN \cite{LSTM-FCN} model features an LSTM layer and stacked CNN layers which directly extract features from time series. These features are subsequently fed into a softmax layer to produce class probabilities. However, it has a limitation in capturing long dependencies among different variables. To address this, Hao et al \cite{CA-SFCN}. proposed to use of two cross-attention modules to enhance their CNN-based model. TapNet \cite{tapnet} constructs an attentional prototype network that incorporates LSTM, and CNN to learn multi-dimensional interaction features. RLPAM \cite{RLPAM} adopts a unique approach by transforming MTS into a univariate cluster sequence and subsequently employs reinforcement learning for pattern selection. WHEN \cite{when} was proposed to learn heterogeneity by utilising a hybrid attention network, incorporating both DTW attention and wavelet attention.

\subsection{Transformer-Based Time Series Classifiers}

In 2017, Vaswani et al. introduced the Transformer architecture, achieving a breakthrough in Natural Language Processing \cite{transformer} and demonstrating notable success in Computer Vision tasks \cite{vision-transformer}. Recently, it has proven effective in time series classification tasks. Specifically, GTN \cite{gatedtrans} utilises a two-tower multi-headed attention approach for extracting distinctive information from the input series. The integration of the output from the two towers is achieved through gating, implemented by a learnable matrix. ConvTran \cite{ConvTran} was proposed to enhance the position embedding by leveraging both absolute and relative position encoding. SVP-T \cite{svp-t} uses clustering to identify time series subsequences and employs them as inputs for the Transformer, enabling the capture of long- and short-term dependencies among subseries. Recently, the application of pretrained transformer-based self-supervised learning models like BERT \cite{bert} has achieved significant success not only in the field of NLP but also in other areas \cite{text-da,nayer,text-graph,lander}. Inspired by these successes, many models attempt to adopt a similar structure for time series classification \cite{TST, ts2vec}. It is noteworthy that most previous transformer-based methods effectively exploit the generic information of time series.

\subsection{Shapelet Discovery for Time Series}

Shapelets refer to short subsequences within time series that contain class-specific information by exhibiting a small distance to the time series of the target class and a larger distance to other classes (see Figure \ref{fig:shapelets}). Additionally, each shapelet can encompass crucial subsequences located at different positions and variables within a time series. This coverage enables them to effectively represent the time series. In the last decade, the effectiveness of shapelets for time series has been proven by many related studies \cite{ppsn, shapelet, st, ls, shapenet}. The original shapelet discovery method \cite{shapelet} extracts all possible subsequences in the training set and considers the subsequences as shapelets when they have the highest information gain ratio. It requires excessive computing time and is hard to apply to MTSC. Other methods use random shapelets that lack position and variable information \cite{fastshapelet-mtc}, or employ the common subsequences as shapelets, which unfortunately have limited discriminative features \cite{shapenet}. Recently, \cite{ppsn} proposed the hyperfast Offline Shapelet Discovery (OSD), which utilises important points to extract a small number of high-quality shapelets from the original time series data. It has been demonstrated to be a SOTA method for univariate time series classification.

\section{Preliminaries}

\noindent
\textbf{Multivariate Time Series Classification.} We represent MTS as $\mathbf{X} \in \mathbb{R}^{V \times T}$, where $V$ denotes the number of variables and $T$ represents the length of the time series. Here, $\mathbf{X} = {X^1, \ldots, X^V}$, and each $X^v$ corresponds to a time series for variable $v$. Specifically, $X^v = {x^v_1, \ldots, x^v_T}$, where $x^v_1$ signifies a value for variable $v$ at timestamp $t$ within $\mathbf{X}$. Consider a training dataset $\mathcal{D} =\{ (\mX_i, \mY_i)\}_{i=1}^M$, where $M$ is the number of time series instances and the pair $(\mX_i, \mY_i)$ represents a training sample and its corresponding label, respectively. The objective of MTSC is to train a classifier $f(X)$ to predict a class label for a multivariate time series with an unknown label.

\noindent
\textbf{Time Series Subsequence.} Given a time series $X$ of length $T$, a time series subsequence $X[p_s:p_e] = x_{p_s},..., x_{p_e}$ is a consecutive subsequence of time series $X$, where $p_s$ is a start index and $p_e$ is an end index.

\noindent
\textbf{Perceptual Subsequence Distance (PSD).} Given a time series $X$ of length $T$, and a subsequence $S = s_1,..., s_l$ of length $l$, with $l \leq T$, the PSD \cite{ppsn} of $X$ and $S$ is determined as:
\begin{equation}
    PSD(X, S) = \min_{j=1}^{T-l+1}\left( \text{CID}(T[j:j+l-1], S) \right)\;,
    \label{eq:psd}
\end{equation}
where CID is the complexity-invariant distance, which is commonly used in time series mining, in general, \cite{cid} and shapelet discovery in particular \cite{ppsn}. 

\begin{figure}[t]
\begin{center}
\includegraphics[width=\linewidth]{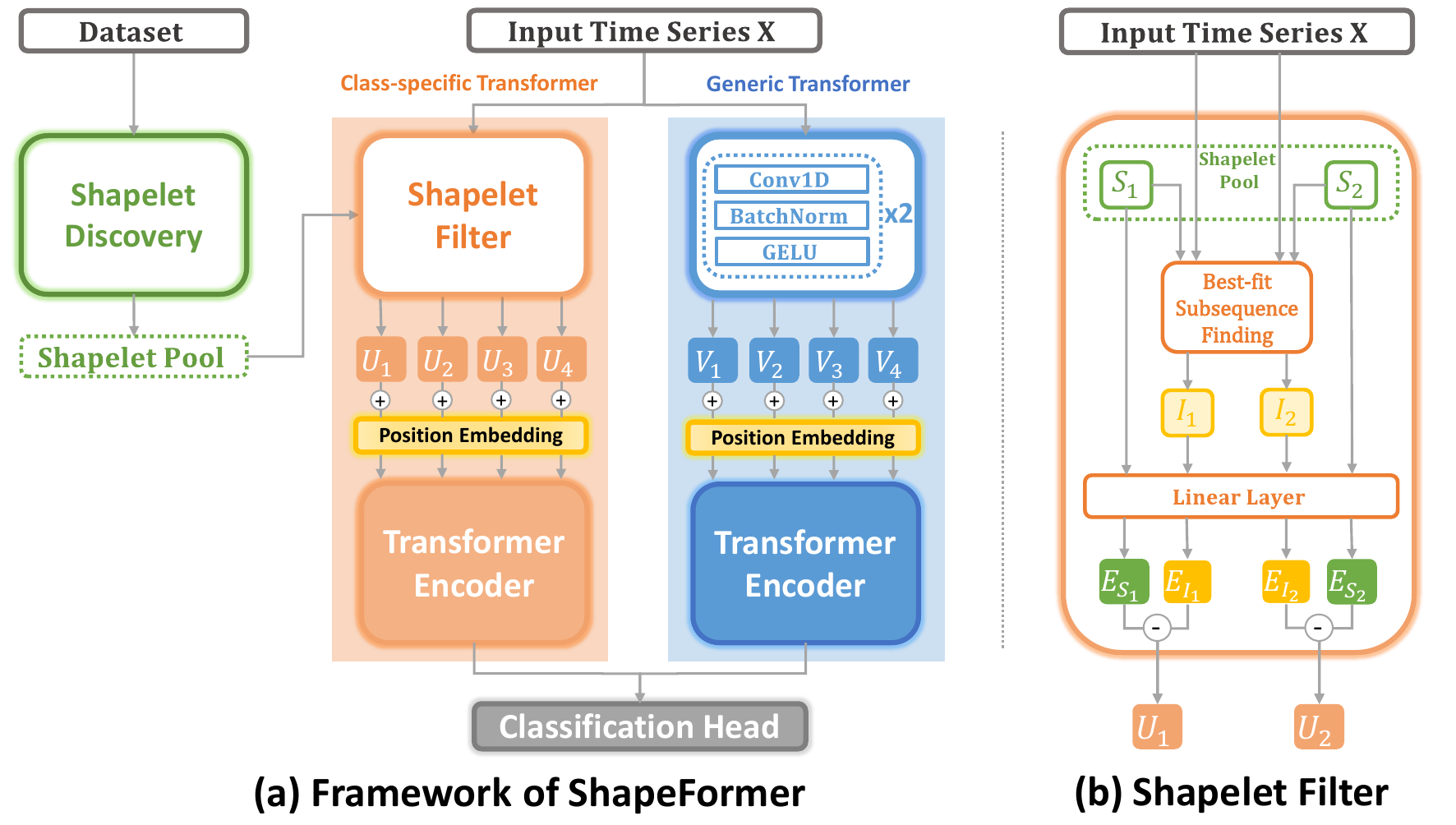}
\end{center}
\caption{The general architecture of ShapeFormer.}
\label{fig:arch}
\end{figure}

\section{Shapelet Transformer Model}

We propose ShapeFormer, a transformer-based method that leverages the strength of both class-specific and generic features in time series. In contrast to existing transformer-based MTSC methods \cite{gatedtrans, TST, svp-t}, our approach first extracts shapelets from the training datasets (Section \ref{sec:se}). Subsequently, these extracted shapelets are used to discover discriminative features in time series through the use of a class-specific transformer module (Section \ref{sec:gt}). Additionally, we introduce the use of convolution layers with a generic transformer module to extract generic features in time series (Section \ref{sec:lt}). Finally, the overall architecture of ShapeFormer is summarised in Section \ref{sec:overall} and Figure \ref{fig:arch}.

\subsection{Shapelet Discovery}
\label{sec:se}
\begin{figure}[t]
\begin{center}
\includegraphics[width=\linewidth]{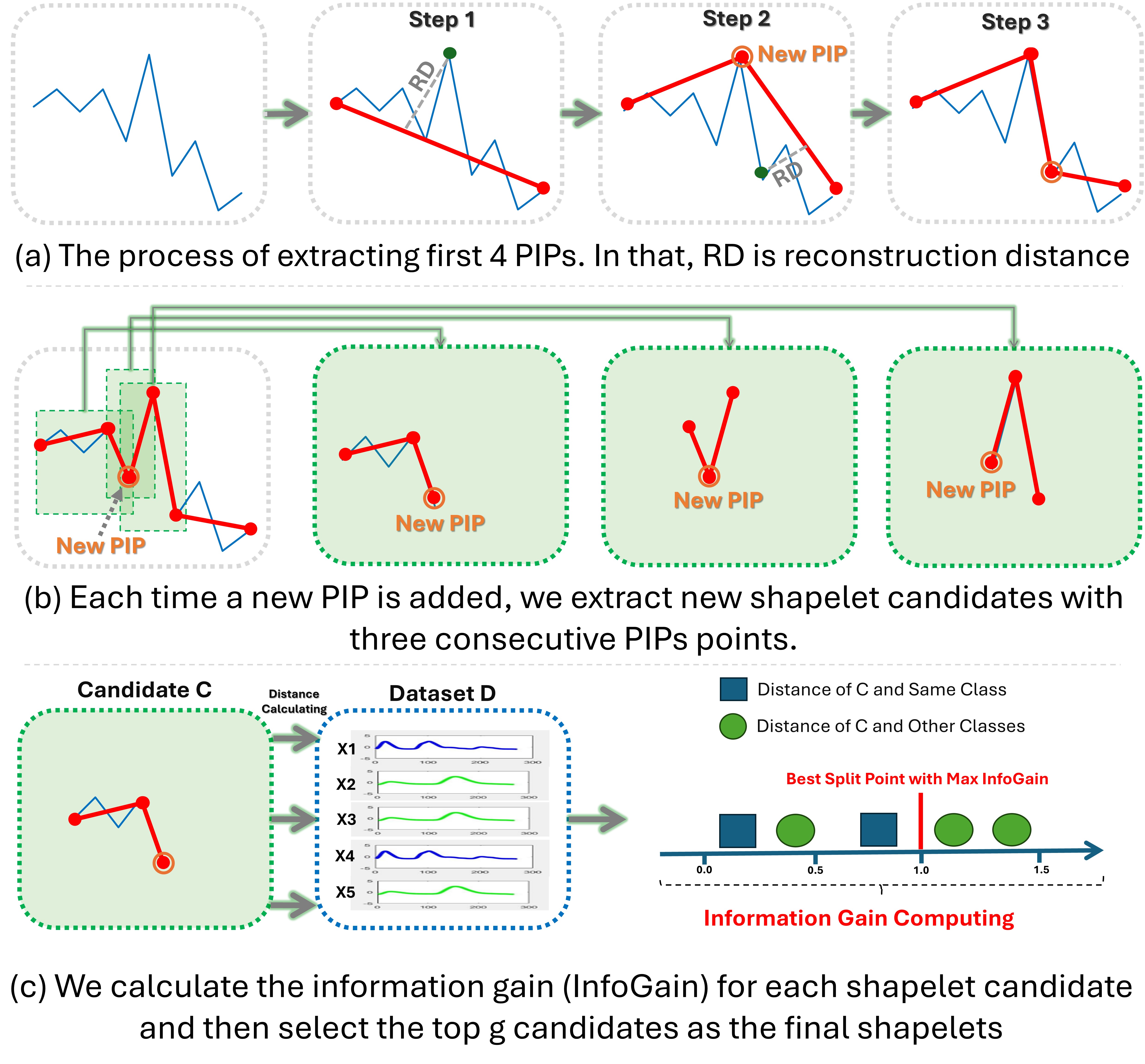}
\end{center}
\caption{The process of Offline Shapelet Discovery.}
\label{fig:osd}
\end{figure}

This section introduces the Offline Shapelet Discovery (OSD) method, inspired by \cite{ppsn}, to multivariate time series. In contrast with other methods, our OSD employs Perceptually Important Points (PIPs) \cite{pip}, condensing time series data by choosing points that closely resemble the original, to efficiently select high-quality shapelets. The selection process is based on the reconstruction distance, with the highest index continuously chosen. We define the reconstruction distance as the perpendicular distance between a target point and a line reconstructed by the two nearest selected important points\cite{ppsn,pip}. The process of our OSD is illustrated in Figure \ref{fig:osd} and the pseudo-code is presented in Algorithm \ref{alg:opse}. Given the dataset $\mathcal{D} =\{ (\mX_i, \mY_i)\}_{i=1}^M$, our method contains two main phases, including shapelet extraction and shapelet selection.

In the first phase, our OSD initially extracts shapelet candidates by identifying PIPs. Specifically, the first and last indices are added to the PIPs set. Subsequently, the index with the highest reconstruction distance is continuously added to the PIPs set. Each time a new PIP is added, we extract new shapelet candidates with three consecutive PIPs points. This means that, with each new PIP, a maximum of three shapelet candidates can be added to the set. In this paper, we set the number of PIPs as $npip = 0.2 \times T$, where $T$ represents the time series length. Our method aims to select a maximum of $3 \times npip$ candidates, therefore, we only extract an average of 5900 candidates for each dataset. This count is significantly smaller than the 45 million candidates typically extracted through classic shapelet discovery methods \cite{shapelet, st}, thereby significantly speeding up the process. We then store four types of information for each shapelet, including the value vector of shapelets, its start index, end index, and variables.

In the second phase, our method selects an equal number of shapelets for each class. Given the shapelet candidate $S_i$ of class $Y_i$, we first compute its PSD with all instances in the training datasets (Eq. \ref{eq:psd}). After that, their distance will be used to find optimal information gain. This implies that the optimal information gain is the highest ratio achievable by the shapelet $S_i$ \cite{ppsn}. Finally, the top $g$ candidates with the highest information gain are chosen as the shapelets and stored in the shapelet pool $\mS$.

\begin{algorithm}[t]
\caption{Offline Shapelet Discovery}
\label{alg:opse}
\kwInput{$\mathcal{D} =\{ (\mX_i, \mY_i)\}_{i=1}^M$: dataset; time series length: $T$; number of variables: $V$, number of PIPs: $k$; number of shapelets: $g$; set of classes: $\mathcal{Y}$; $|Y|$ is the number of classes. }
\Comment{Shapelet Extracting}
$\mathcal{C}$ = []\; 
\ForEach{$X \sim \mD$}{
    \For{$v$ = 1 to $V$}{
        $\mP$ = [1, $T$] \# \textit{Add the first and last index into PIPs set: $P$}\;
        \For{$j$ = 1 to k -2}{
            Find index $p$ from 1 to $T$ with maximum reconstruction distance\;
            $\mP$.append($p$).sorted() \# \textit{Add a new index $p$ into $\mP$}\;
            $idx$ = $\mP$.index($p$)
            \For{$z$ =  0 to 2}{
                \# \textit{Validating newly generated candidates}\;
                \If{$idx+2 \leq |\mI|$ and $idx-z \geq 1$}{
                     $idx_s = \mP[idx-z]$\;
                     $idx_e = \mP[idx+2-z]$\;
                     $C = X[idx_s:idx_e]$\;
                     Add new candidates $C$, its start index $idx_s$, end index $idx_e$, and its variables $v$ into $\mathcal{C}$.
                }
            }        
        }
    }
}
\Comment{Shapelet Selecting}
\ForEach{$S_i \sim \mathcal{C}$}{
    $D$ = []\;
    \ForEach{$X \sim \mD$}{
        $d = \text{PSD}(X, S_i)$ (Eq. \ref{eq:psd})\;
        $D$.append($d$)\;
    }
    Compute the optimal information gain of $S_i$ using $D$\;
}
$\mathcal{S}$ = []\;
\ForEach{$Y^i \sim \mathcal{Y}$}{
    Select the top $g/|Y|$ shapelets ranked by information gain in class $Y^i$ from $\mathcal{C}$;
    Add them to set $\mathcal{S}$;
}
\textbf{return} $\mathcal{S}$
\end{algorithm}

\subsection{Class-Specific Transformer}
\label{sec:gt}

\begin{figure}[t]
\begin{center}
\includegraphics[width=\linewidth]{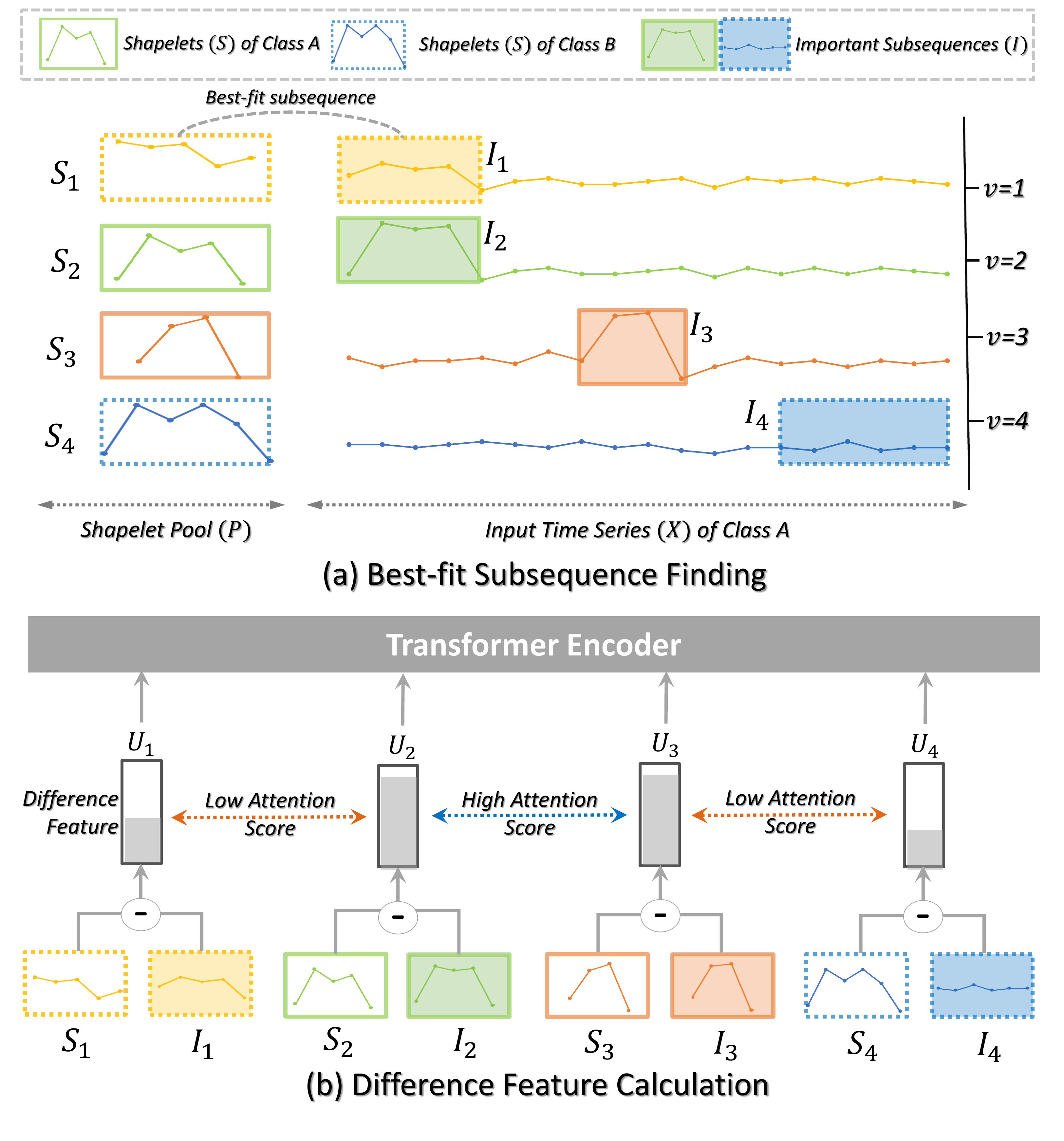}
\end{center}
\caption{The illustrations for: (a) best-fit subsequence finding method; (b) difference feature calculation method.}
\label{fig:moti}
\end{figure}

To utilise the class-specific characteristics of shapelets, we first propose the Shapelet Filter which is used to effectively discover input tokens for the transformer model. 

\noindent
\textbf{Shapelet Filter}. Given a shapelet pool $\mathcal{S}$ (as discussed in Section \ref{sec:se}), an input time series $X$ and its label $Y$, we first select the best-fit subsequence for each shapelet in $\mathcal{S}$ (refer to Figure \ref{fig:moti}a). Specifically, with each shapelet $S_i \in \mathcal{S}$, its length $l$, start index $p^i_s$, end index $p^i_e$ and variables $v^i$, we calculate the distance CID of them with all subsequences in time series $X$ \cite{ppsn}. After that, the subsequence with the shortest distance will be selected as an important subsequence $I_i$ of $S_i$.
\begin{align}
    \text{index} &= \text{argmin}_{j=0}^{T-l+1}{CID(X[j:j+l], S_i)}\;,\\
    I_i &= X[\text{index}:\text{index}+l]\;.
\end{align}
To reduce computing time and effectively utilise the position information of the shapelet, we propose limiting the search for the best-fit subsequence to a neighbouring area within the hyperparameter window size $w$ on both the left and right sides of the actual position of the shapelet. This means that one shapelet only calculates the distance with maximum $2w+1$ subsequences in $X$. 
\begin{align}
    \text{index} &= \text{argmin}_{j=p_s-w}^{p_s+w+1}{CID(X[j:j+l], S_i)}\;,\\
    I_i &= X[\text{index}:\text{index}+l]\;.
\end{align}
Subsequently, we compute the difference features $U_i \in \mathbb{R}^{1\times d_{spe}}$ between the embedding of each shapelet and their most fitting subsequences derived from the input time series (see Figure \ref{fig:moti}b).
\begin{align}
    U_i = \mathcal{P}_I(I_i) - \mathcal{P}_S(S_i)\;,
    \label{eq:u}
\end{align}
where $\mathcal{P}$ is the linear projector of $\mathbb{R}^{l \times d_{spe}}$ with $l$ is length of shapelet and $d_{spe}$ is the embedding size of difference features. 

Similar to the distance between shapelet and time series, our difference feature also highlights the substantial distinctions among classes. Furthermore, by directly incorporating the shapelets in computing the difference features (Eq. \ref{eq:u}), the shapelets are now considered as the learnable parameters of the Shapelet Filter component. Therefore, rather than using fixed shapelets, we can use them as the initial parameters of the Shapelet Filter, which will be optimised during training.

\noindent
\textbf{Position Embedding}. The difference features $ U_i $ are then integrated with position embeddings to capture their order. To better indicate the position information of shapelets, the embeddings of three types of positions are considered, including the start index, end index, and variables. Specifically, we propose to use a one-hot vector representation for these indices and then employ a linear projector to learn their embedding.
\begin{align}
    \text{PE}(p) = \code{Linear}(\text{one-hot}(p))\;,\\
    U_i = U_i + \text{PE}(p_s^i) + \text{PE}(p_e^i) + \text{PE}(v^i)\;.
    \label{eq:pos}
\end{align}
We also observed that the performance is enhanced when we only use the position of shapelets instead of the position of best-fit subsequences. This improvement can be attributed to the fact that the fixed position is easier to learn than the unstable position of best-fit subsequences.

\noindent
\textbf{Transformer Encoder}. The class-specific difference features, along with their corresponding position embeddings, are then input into a transformer encoder to learn their correlation. Specifically, we employ the multi-head attention mechanism (MHA) \cite{transformer} for this purpose.  Given an input series, $\mU = U_1,\ldots, U_g$ and the projections $W_q$, $W_k$, $W_v \in \mathbb{R}^{d_{spe} \times d_{spe}}$ represent query, key, and value matrices, respectively. These matrices, $W_q$, $W_k$, $W_v$, undergo reshaping into $\mathbb{R}^{h \times d_{spe} \times (d_{spe}/h)}$ to signify the $h$ attention heads and are subsequently concatenated into standard dimensions after computation. Each attention head within this set is capable of capturing distinct relationships of the features. Finally, these matrices are used to compute an output $\mathbf{Z}^{\text{spe}} = Z_1^{\text{spe}}, ..., Z_g^{\text{spe}}$ where $Z_i^{\text{spe}} \in \mathbb{R}^{d_{spe}}$:
\begin{align}
    Z_i^{\text{spe}} = \sum_{j=1}^{g}{a_{i,j}(U_j*W_v)}\;,
\end{align}
where $a_{i,j}$ is an attention score which is calculated as:
\begin{align}
    a_{i,j} &= \code{softmax}(\frac{(U_i*W_q)(U_j*W_k)}{\sqrt{d_{spe}}})\;,
\end{align}
Thanks to the class-represented characteristics of these features, the attention score for features within the same class is boosted compared to features in different classes. This enhancement helps the model better distinguish between different classes. Additionally, owing to the nature of shapelets, the difference features possess the ability to identify significant subsequences across different temporal locations and variables within the time series. This capability enables the module to effectively capture temporal and variable dependencies in time series data.

\noindent
\textbf{Class Token}. Existing transformer-based methods apply averaging pooling to $\mathbf{Z}^{\text{spe}}$, to obtain the final token for classification \cite{ConvTran, gatedtrans}. However, our class-specific transformer module utilises difference features that capture the distinctive characteristics of each shapelet. Applying average pooling may diminish these properties, potentially limiting performance. To address this, we propose using only the first difference feature of the highest information gain shapelet $Z^{\text{spe}}_1$ as the class token $Z^{\text{spe}}_*$ for final classification. The reason for this is that when averaging all tokens, there is a loss of information regarding distinct features $U_i$. Moreover, the first token $Z_1^{\text{spe}}$, which carries the highest information gain, harbors the most crucial features for effectively classifying time series.

\subsection{Generic Transformer}
\label{sec:lt}
Besides leveraging the power of class-specific features, in this section, we introduce the generic transformer module, utilising convolution filters to extract generic features in the time series. Specifically, we employ two CNN components \cite{ConvTran,djcnn}, each comprising Conv1D, BatchNorm, and GELU, to effectively discover generic features. The first block is designed to capture the temporal patterns in time series by using the Conv1D filter $\in \mathbb{R}^{1\times d_c}$. On the other hand, the second block uses the Conv1D filter $\in \mathbb{R}^{V\times 1}$ to capture the correlation between variables in time series. In this context, $V$ represents the number of variables, and $d_c$ is the kernel size of the convolution filter, which is fixed at 8 in all experiments. From that, the output generic feature $V_i \in \mathbb{R}^{1\times d_{gen}}$ is calculated as follows:
\begin{align}
    \code{ConvBlock(X)} &= \code{GELU}(\code{BatchNorm}(\code{Conv1D}(X)))\;,\\
    \mV &= \code{ConvBlock}(\code{ConvBlock}(X))\;.
\end{align}
Afterward, these features will be fed to the $h$ multi-attention heads to learn the correlation. Each attention head has the capacity to capture distinct patterns within time series data.
\begin{align}
    \mZ^{\text{gen}} = \text{MHA}(\mV + \mP)\;,
\end{align}
as the position of each element $V_i \in \mV$ lacks inherent meaning, we utilise the learnable position embedding $\mP$ for representing them. Furthermore, since the module takes classic features as input tokens, we employ averaging pooling to derive the final class token.
\begin{align}
    \mZ^{\text{gen}}_* = \code{AvgPooling}(\mZ^{\text{gen}})\;.
\end{align}

\subsection{Overall Architecture of ShapeFormer}
\label{sec:overall}

To enhance clarity, we present the overall architecture of ShapeFormer in Figure \ref{fig:arch}. Our method initiates by extracting shapelets from the training datasets. Subsequently, for a given input time series $\mX$, it is processed through dual transformer modules, comprising the class-specific shapelet transformer and the generic convolution transformer. The outputs from these two modules are then concatenated and fed into the final classification head.
\begin{align}
    \mZ &= \code{concat}(\mZ^{\text{spe}}_*, \mZ^{\text{gen}}_*)\;,\\
    \hat{y} &= \code{argmax}(\code{softmax}(\code{Linear}(\mZ)))\;.
\end{align}
The predictions $\hat{y}$ are used to optimise the model parameters based on the following objective function:
\begin{align}
    \mathcal{L} &= \mathcal{L}_{\text{CE}}(\hat{y}, Y)\;.
\end{align}
where, CE is the Cross-Entropy Loss, which can be calculated as $\mathcal{L}_{\text{CE}}(\hat{y}, Y) = \sum_i^{|Y|}{y_ilog(\hat{y_i})}$.

\begin{table*}[t]
\caption{Accuracies of our proposed method ShapeFormer and 12 compared methods on all datasets of the UEA archive \cite{uea}.}
\begin{adjustbox}{width=\linewidth}
\begin{tabular}{l|ccccccccccccc}
\toprule
                          & EDI            & DTWD           & \begin{tabular}[c]{@{}c@{}}WEASEL\\      +MUSE\end{tabular} & MiniRocket     & LCEM           & \begin{tabular}[c]{@{}c@{}}MLSTM\\      -FCNs\end{tabular} & Tapnet         & Shapenet       & WHEN           & TST            & ConvTran       & SVPT           & Our            \\
\midrule
ArticularyWordRecognition & 0.970          & 0.987          & 0.990                                                       & 0.992          & \textbf{0.993} & 0.973                                                      & 0.987          & 0.987          & \textbf{0.993} & 0.983          & 0.983          & \textbf{0.993} & \textbf{0.993} \\
AtrialFibrillation        & 0.267          & 0.220          & 0.333                                                       & 0.133          & 0.467          & 0.267                                                      & 0.333          & 0.400          & 0.467          & 0.200          & 0.400          & 0.400          & \textbf{0.660} \\
BasicMotions              & 0.676          & 0.975          & \textbf{1.000}                                              & \textbf{1.000} & \textbf{1.000} & 0.950                                                      & \textbf{1.000} & \textbf{1.000} & \textbf{1.000} & 0.975          & \textbf{1.000} & \textbf{1.000} & \textbf{1.000} \\
CharacterTrajectories     & 0.964          & 0.989          & 0.990                                                       & 0.993          & 0.979          & 0.985                                                      & \textbf{0.997} & 0.980          & 0.996          & 0.000          & 0.992          & 0.990          & 0.996          \\
Cricket                   & 0.944          & \textbf{1.000} & \textbf{1.000}                                              & 0.986          & 0.986          & 0.917                                                      & 0.958          & 0.986          & \textbf{1.000} & 0.958          & \textbf{1.000} & \textbf{1.000} & \textbf{1.000} \\
DuckDuckGeese             & 0.275          & 0.600          & 0.575                                                       & 0.650          & 0.375          & 0.675                                                      & 0.575          & \textbf{0.725} & 0.700          & 0.480          & 0.620          & 0.700          & \textbf{0.725} \\
ERing                     & 0.133          & 0.929          & 0.133                                                       & \textbf{0.981} & 0.200          & 0.133                                                      & 0.133          & 0.133          & 0.959          & 0.933          & 0.963          & 0.937          & 0.966          \\
EigenWorms                & 0.549          & 0.618          & 0.890                                                       & \textbf{0.962} & 0.527          & 0.504                                                      & 0.489          & 0.878          & 0.893          & N/A            & 0.593          & 0.925          & 0.925          \\
Epilepsy                  & 0.666          & 0.964          & \textbf{1.000}                                              & \textbf{1.000} & 0.986          & 0.761                                                      & 0.971          & 0.987          & 0.993          & 0.920          & 0.986          & 0.986          & 0.993          \\
EthanolConcentration      & 0.293          & 0.323          & 0.430                                                       & \textbf{0.468} & 0.372          & 0.373                                                      & 0.323          & 0.312          & 0.422          & 0.337          & 0.361          & 0.331          & 0.378          \\
FaceDetection             & 0.519          & 0.529          & 0.545                                                       & 0.620          & 0.614          & 0.545                                                      & 0.556          & 0.602          & 0.658          & \textbf{0.681} & 0.672          & 0.512          & 0.658          \\
FingerMovements           & 0.550          & 0.530          & 0.490                                                       & 0.550          & 0.590          & 0.580                                                      & 0.530          & 0.580          & 0.660          & \textbf{0.776} & 0.560          & 0.600          & 0.700          \\
HandMovementDirection     & 0.278          & 0.231          & 0.365                                                       & 0.392          & \textbf{0.649} & 0.365                                                      & 0.378          & 0.338          & 0.554          & 0.608          & 0.405          & 0.392          & 0.486          \\
Handwriting               & 0.200          & 0.286          & \textbf{0.605}                                              & 0.507          & 0.287          & 0.286                                                      & 0.357          & 0.452          & 0.561          & 0.305          & 0.375          & 0.433          & 0.507          \\
Heartbeat                 & 0.619          & 0.717          & 0.727                                                       & 0.771          & 0.761          & 0.663                                                      & 0.751          & 0.756          & 0.780          & 0.712          & 0.785          & 0.790          & \textbf{0.800} \\
InsectWingbeat            & 0.128          & N/A            & N/A                                                         & 0.595          & 0.228          & 0.167                                                      & 0.208          & 0.250          & 0.657          & 0.684          & \textbf{0.713} & 0.184          & 0.314          \\
JapaneseVowels            & 0.924          & 0.949          & 0.973                                                       & 0.989          & 0.978          & 0.976                                                      & 0.965          & 0.984          & 0.995          & 0.994          & 0.989          & 0.978          & \textbf{0.997} \\
LSST                      & 0.456          & 0.551          & 0.590                                                       & 0.643          & 0.652          & 0.373                                                      & 0.568          & 0.590          & 0.663          & 0.381          & 0.616          & 0.666 & \textbf{0.700} \\
Libras                    & 0.833          & 0.870          & 0.878                                                       & 0.922          & 0.772          & 0.856                                                      & 0.850          & 0.856          & 0.933          & 0.844          & 0.928          & 0.883          & \textbf{0.961} \\
MotorImagery              & 0.510          & 0.500          & 0.500                                                       & 0.550          & 0.600          & 0.510                                                      & 0.590          & 0.610          & 0.630          & N/A            & 0.560          & 0.650          & \textbf{0.670} \\
NATOPS                    & 0.850          & 0.883          & 0.870                                                       & 0.928          & 0.916          & 0.889                                                      & 0.939          & 0.883          & 0.978          & 0.900          & 0.944          & 0.906          & \textbf{0.989} \\
PEMS-SF                   & \textbf{0.973} & 0.711          & N/A                                                         & 0.522          & 0.942          & 0.699                                                      & 0.751          & 0.751          & 0.925          & 0.919          & 0.828          & 0.867          & 0.925          \\
PenDigits                 & 0.705          & 0.977          & 0.948                                                       & N/A            & 0.977          & 0.978                                                      & 0.980          & 0.977          & 0.987          & 0.974          & 0.987          & 0.983          & \textbf{0.990} \\
PhonemeSpectra            & 0.104          & 0.151          & 0.190                                                       & 0.292          & 0.288          & 0.110                                                      & 0.175          & 0.298          & 0.293          & 0.088          & \textbf{0.306} & 0.176          & 0.293          \\
RacketSports              & 0.868          & 0.803          & 0.934                                                       & 0.868          & \textbf{0.941} & 0.803                                                      & 0.868          & 0.882          & 0.934          & 0.829          & 0.862          & 0.842          & 0.895          \\
SelfRegulationSCP1        & 0.771          & 0.775          & 0.710                                                       & \textbf{0.925} & 0.839          & 0.874                                                      & 0.652          & 0.782          & 0.908          & \textbf{0.925} & 0.918          & 0.884          & 0.911          \\
SelfRegulationSCP2        & 0.483          & 0.539          & 0.460                                                       & 0.522          & 0.550          & 0.472                                                      & 0.550          & 0.578          & 0.589          & 0.589          & 0.583          & 0.600          & \textbf{0.633} \\
SpokenArabicDigits        & 0.967          & 0.963          & 0.982                                                       & 0.620          & 0.973          & 0.990                                                      & 0.983          & 0.975          & \textbf{0.997} & 0.993          & N/A            & 0.986          & \textbf{0.997} \\
StandWalkJump             & 0.200          & 0.200          & 0.333                                                       & 0.333          & 0.400          & 0.067                                                      & 0.400          & 0.533          & 0.533          & 0.267          & 0.333          & 0.467          & \textbf{0.600} \\
UWaveGestureLibrary       & 0.881          & 0.903          & 0.916                                                       & 0.938          & 0.897          & 0.891                                                      & 0.894          & 0.906          & 0.919          & 0.903          & 0.891          & \textbf{0.941} & 0.922          \\
\midrule
Average rank              & 11.200         & 9.783          & 7.933                                                       & 5.900          & 6.600          & 9.833                                                      & 8.233          & 6.850          & 3.117          & 8.117          & 5.650          & 5.283          &\textbf{ 2.500}          \\

Number of top-1                & 1              & 1              & 4                                                           & 6              & 4              & 0                                                          & 2              & 2              & 4              & 3              & 4              & 5              & \textbf{15}             \\
Wins                      & 29             & 29             & 24                                                          & 21             & 25             & 30                                                         & 28             & 27             & 16             & 25             & 24             & 24             & -              \\
Draws                     & 0              & 1              & 2                                                           & 2              & 2              & 0                                                          & 1              & 2              & 9              & 0              & 2              & 5              & -              \\
Loses                     & 1              & 0              & 4                                                           & 7              & 3              & 0                                                          & 1              & 1              & 5              & 5              & 4              & 1              & -        \\

P-value & 0.000	& 0.000 &	0.001&0.014&	0.003&	0.000&	0.000&	0.002&	0.475&	0.005&	0.024&	0.000 & - \\
\bottomrule
\end{tabular}
\end{adjustbox}
\label{tab:result}
\end{table*}

\section{Experiments}

\subsection{Experimental Setting}

\noindent
\textbf{Dataset.} We assess our approach using the UEA archive, a well-known benchmark made up of 30 distinct datasets for MTSC \cite{uea}. It covers various domains, including Human Activity Recognition, Motion classification, ECG classification, EEG/MEG classification, Audio Spectra classification, and more. The sample sizes of datasets in the UEA archive range from 27 to 50,000, the time series lengths spanning 8 to 17,984, and dimensions varying from 2 to 1,345.

\noindent
\textbf{Metrics.} We use classification accuracy to evaluate model performance and compare methods based on their average ranks and win/draw/loss counts on all datasets. Finally, we evaluate the statistical significance of performance differences using the p-value of Friedman and Wilcoxon signed-rank test \cite{uea}.


\noindent
\textbf{Implementation Details.} Our model was trained using the RAdam optimiser with an initial learning rate set as 0.01, a momentum of 0.9, and a weight decay of 5e-4. The training process involved a batch size of 16 for a total of 200 epochs. We configured the number of attention heads to be 16 and followed the protocol outlined in \cite{svp-t, TST}. This protocol involves splitting the training set into 80\% for training and 20\% for validation, allowing us to fine-tune hyperparameters. Once the hyperparameters were finalised, we conducted model training on the entire training set and subsequently evaluated its performance on the designated official test set.

\noindent
\textbf{Environment.} All the experiments are conducted on a machine with one Intel(R) Xeon(R) Silver 4214 CPU @ 2.20GHz and one NVIDIA Tesla V100 SXM2.

\subsection{Baselines} 
We have selected 12 baseline methods for the comparative experiments, comprising two distance-based methods: \textit{EDI}, \textit{$DTW_D$} \cite{uea}; a pattern-based algorithm: \textit{WEASEL+MUSE} \cite{weasel}; a feature-based algorithm: \textit{MiniRocket} \cite{minirocket}; an ensemble method: \textit{LCEM} \cite{lcem}; three deep learning models: \textit{MLSTM-FCNs} \cite{mlstm-fcns}, \textit{Tapnet} \cite{tapnet}, \textit{Shapenet} \cite{shapenet}; an attention-based model: \textit{WHEN} \cite{when}; and three transformer-based models: \textit{TST} \cite{TST}, \textit{ConvTran} \cite{ConvTran}, \textit{SVP-T} \cite{svp-t}. They all attained the SOTA performance described in the most recent research. The details of 12 baseline methods are shown in Appendix \ref{ap:baseline}.

\subsection{Performance Evaluation} 

Table \ref{tab:result} illustrates the experimental results of our method with 12 other competitors on the UEA multivariate time series classification archive \cite{uea}. The accuracy of 12 baseline methods are taken from \cite{svp-t}, except the results of WHEN, and ConvTran are taken from their original papers \cite{when, ConvTran}. The best result on each dataset is indicated in bold, and the summarised information is provided in the last six lines of the table. 

The results show that among all methods, ShapeFormer achieves the best performance in both the highest average rank (2.5) and the largest number of top-1 (the best in 15 out of 30 datasets). This indicates that ShapeFormer can be taken as a SOTA for MTSC. The rank index signifies that, even on some datasets where our model does not exhibit the highest performance, its results remain highly competitive. Specifically, the average rank of our method is slightly higher compared to that of the runner-up, WHEN, a difference of 0.617. Meanwhile, the gap in average rank between ShaperFormer and three Transformer-based methods (TST, ConvTran, SVP-T) is large, with 5.617, 3.15, and 2.783 respectively. The p-value is $ \leq 0.05$, which confirms there ranks have statistically significant differences. Specifically, the p-values for ShapeFormer in comparison to all methods are below 0.05, which indicates the results are statistically significant except for WHEN. However, regarding the number of top-1, our ShapeFormer attained SOTA results in 15 datasets compared to WHEN, only 4 datasets.

\subsection{Ablation Study and Model Design}

\begin{figure}[t]
\begin{center}
\includegraphics[width=0.8\linewidth]{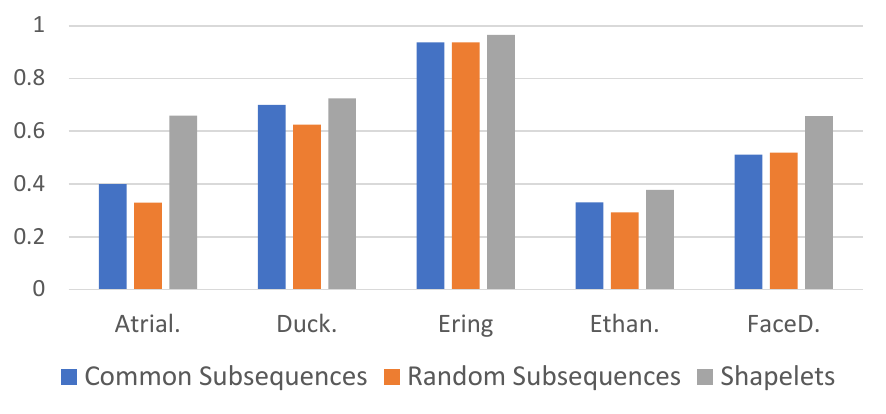}
\end{center}
\caption{Accuracies of using shapelets and two other types of subsequences.}
\label{fig:as-shapelet}
\end{figure}

\noindent
\textbf{Effectiveness of Using Shapelets.} In Figure \ref{fig:as-shapelet}, we compare the performance when using random subsequences, common subsequences as mentioned in \cite{svp-t}, and shapelets in our methods. The results demonstrate that the shapelets outperform the other two methods in terms of accuracy across all five datasets. This highlights the benefit of highly discriminative shapelet features in increasing the performance of the transformer-based model, thereby indicating the contribution of our work. 

\begin{figure}[t]
\begin{center}
\includegraphics[width=0.9\linewidth]{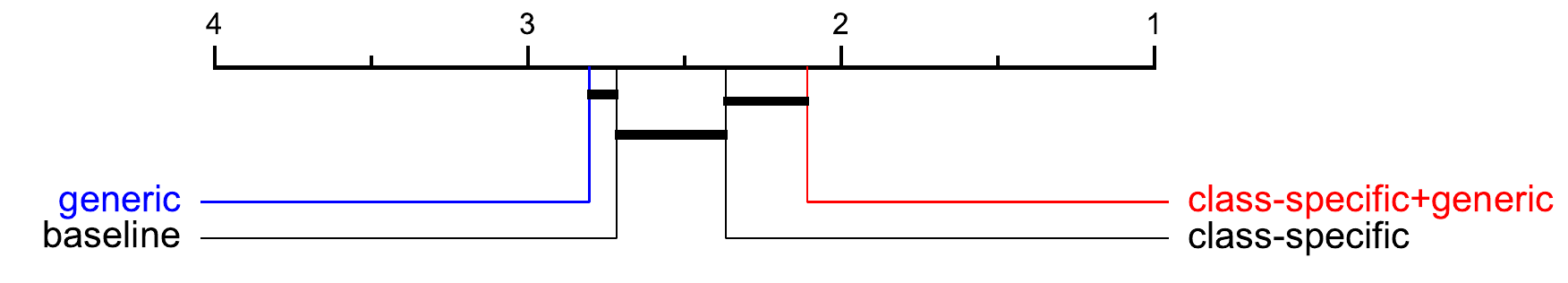}
\end{center}
\caption{Average ranks for 3 variations of ShapeFormer and the baseline (SVP-T \cite{svp-t} - the current SOTA transformer-based method).}
\label{fig:as-component}
\end{figure}

\noindent
\textbf{Component Evaluation.} We begin by evaluating the impact of two key modules in our ShapeFormer: the Class-specific Transformer (Section \ref{sec:gt}) and the Generic Transformer (Section \ref{sec:lt}), in comparison with the baseline method, SVP-T \cite{svp-t} on the first 10 datasets of UEA archive \cite{uea}. In this process, individual components are incrementally incorporated to assess their impact on the ultimate accuracy. As depicted in Figure \ref{fig:as-component}, applying the generic transformer alone exhibits a lower accuracy compared to the baseline. In contrast, utilising only the class-specific module results in significantly improved performance over the baselines, emphasising the effectiveness of class-specific features in the transformer-based time series model. Furthermore, the combination of class-specific and generic components shows a positive impact on the enhancement of classification accuracy. This combination harnesses the power of both features, significantly boosting overall performance.

\begin{figure}[t]
\begin{center}
\includegraphics[width=0.85\linewidth]{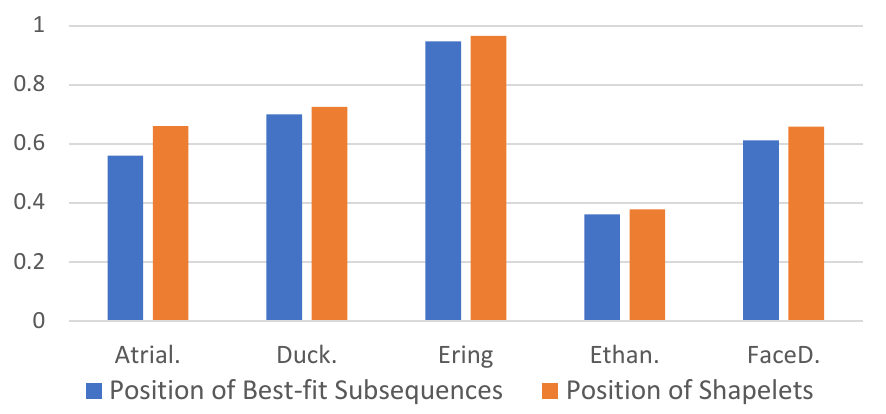}
\end{center}
\caption{Accuracies of using the position of best-fit subsequences and shapelets.}
\label{fig:as-pos}
\end{figure}

\noindent
\textbf{Choosing between the Position of Shapelets and Best-fit Subsequences.} Our ShapeFormer leverages shapelets to find the best-fit subsequences and employ the difference features $U_i$ calculated by them as the inputs for the Transformer encoder. Then there is a question \textit{"Should we choose the positions of the shapelets or the best-fit subsequences for position embedding?"}. Figure \ref{fig:as-pos} illustrates a comparison between the accuracies achieved by employing the position of the best-fit subsequences and shapelets, as indicated in Eq. \ref{eq:pos}, for position embedding in the Transformer encoder. The outcomes indicate that our approach exhibits superior performance when utilising the position of shapelets across all five datasets under consideration. This enhancement can be ascribed to the fact that learning from the fixed position of shapelets is easier compared to the unstable position of the best-fit subsequences.

\begin{figure}[t]
\begin{center}
\includegraphics[width=0.9\linewidth]{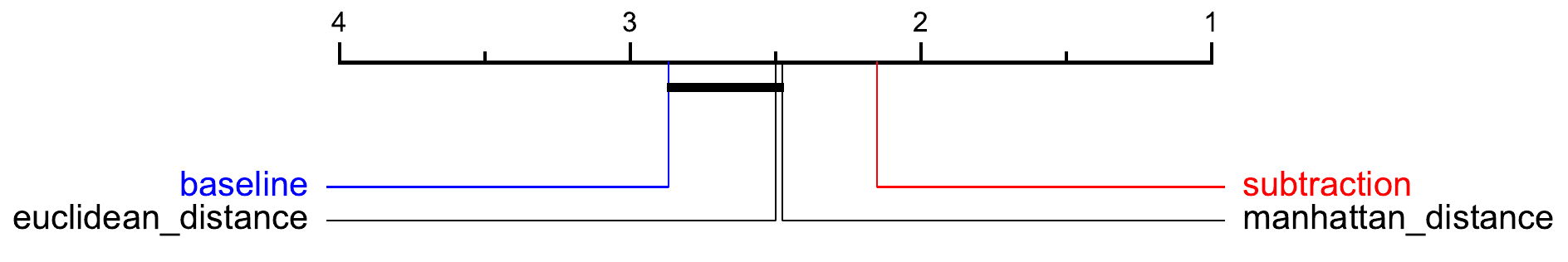}
\end{center}
\caption{Average accuracy ranks of various calculation methods for difference features.}
\label{fig:as-dc}
\end{figure}

\noindent
\textbf{Comparison with Various Methods for Calculating Difference Features.} The critical difference diagram in Figure \ref{fig:as-dc} displays the performance of using different methods for calculating difference features in Eq. \ref{eq:u}, including Manhattan Distance, Euclidean Distance, and the subtraction between $\mathcal{P}_I(I_i)$ and $\mathcal{P}_S(S_i)$. The results demonstrate that: 1) All calculation methods for difference features yield better results compared to the SVP-T baseline; 2) Using subtraction exhibits the highest performance. Although the subtraction is simple, its superiority lies in effectively capturing relative changes by considering both the magnitude and direction of changes between embedding vectors $\mathcal{P}_I(I_i)$ and $\mathcal{P}_S(S_i)$.

\begin{figure}[t]
\begin{center}
\includegraphics[width=0.9\linewidth]{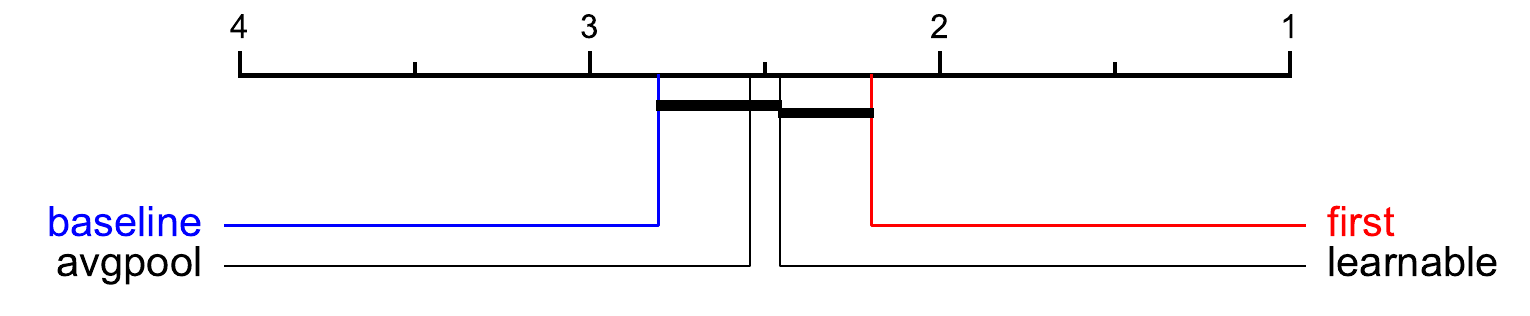}
\end{center}
\caption{Average accuracy ranks of different class token designs.}
\label{fig:as-td}
\end{figure}

\noindent
\textbf{Different Class Token Designs.} The output of the class-specific transformer consists of a series of tokens $Z_1^{\text{spe}},\ldots,Z_g^{\text{spe}}$. The question at hand is, \textit{"Are there any effective ways to design class tokens before feeding them to the classification head?"}. In Figure \ref{fig:as-td}, we analyse the impact of different class token designs on the performance of ShapeFormer. The results indicate that: 1) Our ShapeFormer outperforms the baseline with all types of class token designs, demonstrating the advantage of our method; 2) Utilising the first token $Z_1^{\text{spe}}$ as the final class token $Z_*^{\text{spe}}$ for ShapeFormer yields the best performance. This is due to the fact that learning or averaging all tokens results in a loss of information on difference features $U_i$. Furthermore, the first token, containing the highest information gain, possesses the most discriminative features for classifying time series.

\subsection{Hyperparameter Sensitivity}
\label{sec:hyper}

\noindent
\textbf{Tuning Window Size and Number of Shapelets.} In our method, there are two main parameters related to shapelet discovery that need tuning, including the window size when calculating the distances between shapelets and subsequences and the number of shapelets. Regarding the window size, we propose to tune it exclusively during the shapelet discovery phase. For each dataset, we will select a window size from the set [10, 20, 50, 100, 200], aiming to provide the top 100 shapelets with the highest information gain. This tuning technique can significantly reduce training time since it only operates during the shapelet discovery phase. As for the number of shapelets, considering the diverse characteristics of different datasets, we choose this number from [1, 3, 10, 30, 100]. The details of our tuned parameters are shown in Appendix \ref{sec:ab-para}.

\noindent
\textbf{Number of PIPs.} As shown in the following Table \ref{tab:Npips}, the model accuracy increases as we increase the number of PIPs (npips) from 0.05T to 0.2T. Afterward, accuracy remains stable even with further increases in npips. Therefore, we set npips at 0.2 for all of our experiments. 

\begin{table}[t]
\caption{The average accuracy for various numbers of PIPs.}
\begin{adjustbox}{width=\linewidth}
    \centering
    \begin{tabular}{|c|c|c|c|c|c|c|c|c|}
    \hline
    \textbf{Npips ($\times T$)} & 0.05 & 0.1 & 0.15 & \textbf{0.2} & 0.25 & 0.3 & 0.4 & 0.5\\
    \hline
    \textbf{Accuracy} & 0.832 & 0.848 & 0.856 & \textbf{0.864} & 0.864 & 0.864 & 0.864 & 0.864\\
    \hline
    \end{tabular}
\end{adjustbox}
    \label{tab:Npips}
\end{table}

\begin{figure}[t]
\begin{center}
\includegraphics[width=\linewidth]{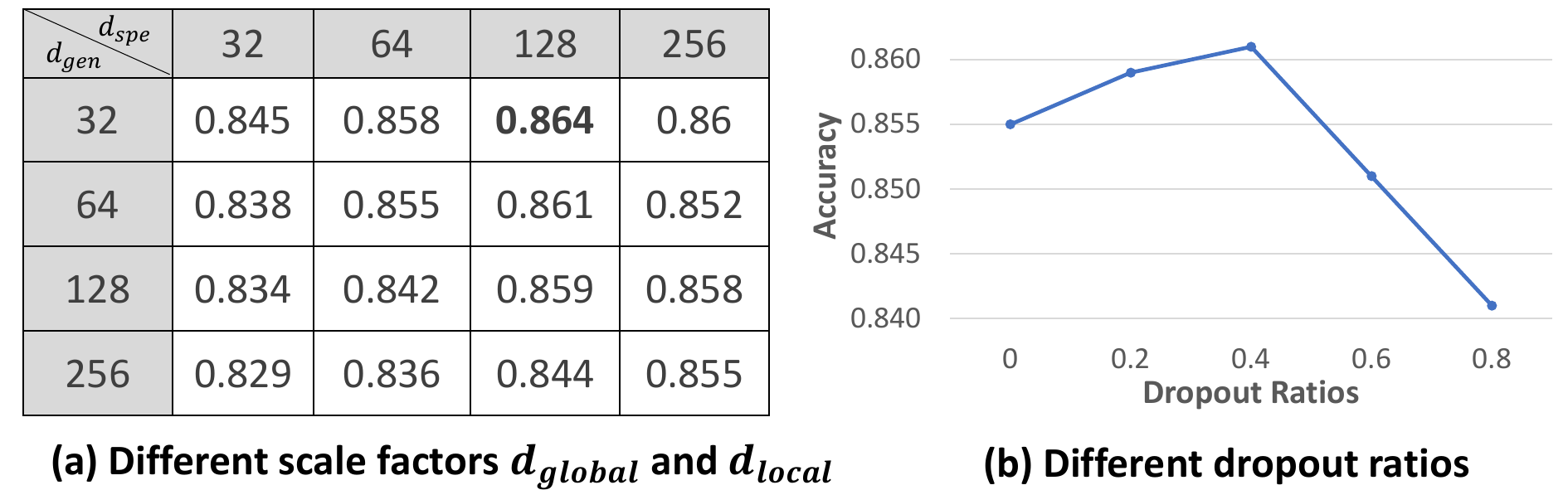}
\end{center}
\caption{Effectiveness of (a) class-specific and generic scale factors and (b) different dropout ratios.}
\label{fig:as-scale}
\end{figure}

\noindent
\textbf{The Scale Factors $d_{\text{spe}}$ and $d_{\text{gen}}$.} In Figure \ref{fig:as-scale}a, we compare the impact of different scale factors of the class-specific and generic embedding sizes on the classification accuracy of ShapeFormer. The results show that: 1) The pair of $d_{\text{spe}} = 128$ and $d_{\text{gen}} = 32$ has achieved the highest accuracy; 2) In general, a larger class-specific embedding size has achieved better performance, indicating the benefit of using shapelets in a transformer-based time series model.

\noindent
\textbf{Dropout Ratios.} In Figure \ref{fig:as-scale}b, we analyse the impact of different dropout ratios of ShapeFormer. It is evident that our methods work well and achieve high performance with small dropout ratios, with the ratio of 0.4 yielding the highest performance.

\subsection{Improving Performance in Specific Datasets by Optimizing Scale Factor}

In MTSC, it is crucial to develop models that generalise well across a majority of datasets rather than models tailored to specific datasets. For example, in terms of \textit{InsectWingbeat} dataset, we observed that setting $d_{\text{gen}}$ (embedding size of generic feature) to 256 leads to significantly better performance (0.704) compared to $d_{\text{gen}}$ at 32 (0.314) (our chosen parameter). However, this improvement comes at the cost of decreased performance on other datasets (from 0.864 to 0.831). Therefore, we recommend tuning this hyperparameter to achieve better performance on specific datasets if needed.

\begin{table}[h]
\caption{Accuracy for InsectWingbeat dataset and the first 10 UEA datasets with various $d_{\text{gen}}$  factors.}
\centering
\begin{adjustbox}{width=\linewidth}
\begin{tabular}{|c|c|c|c|c|}
\hline
\textbf{$d_{\text{gen}}$} & 32 (our choice) & 64 & 128 & 256\\
\hline
\textbf{InsectWingbeat} & 0.314 & 0.500 & 0.634 & \textbf{0.704}\\
\textbf{First 10 UEA datasets} & \textbf{0.864} & 0.852 & 0.844 & 0.831\\
\hline
\end{tabular}
\end{adjustbox}
\label{tab:example}
\end{table}

\subsection{A Case Study of LSST (Imbalanced Datasets)}

\begin{figure}[t]
\begin{center}
\includegraphics[width=\linewidth]{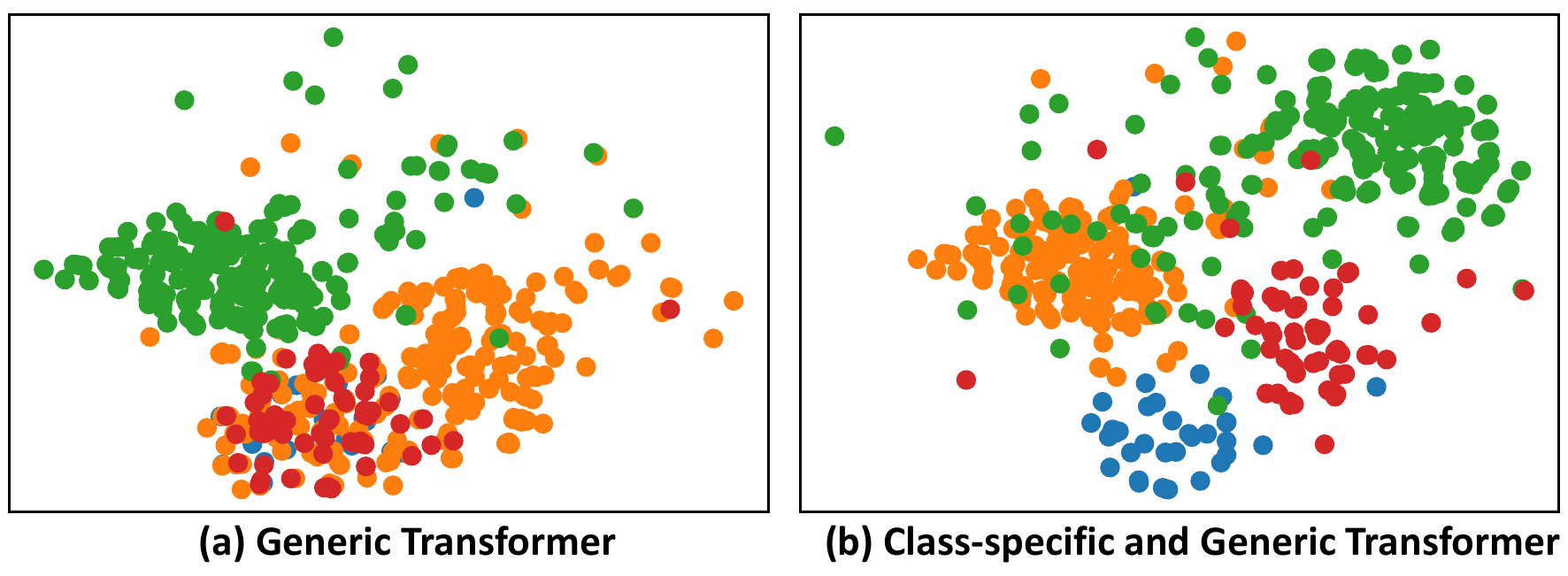}
\end{center}
\caption{The t-SNE visualisation in 4 classes of LSST dataset using (a) our generic transformer and (b) both class-specific and generic transformers. Each point indicates an instance and the colors of the points signify the true labels.}
\label{fig:case1}
\end{figure}

To illustrate the effectiveness of combining both class-specific and generic features transformer modules to classify imbalanced data, we conducted experiments on the \textit{LSST} dataset. The \textit{LSST} dataset comprises 16 classes, and we randomly selected 4 classes to be represented by the colors blue, orange, green, and red, with 35, 270, 382, and 63 instances, respectively. It is clear that the sizes of blue and red classes are significantly smaller compared to the sizes of the green and orange classes. Figure \ref{fig:case1}a shows that the generic transformer prioritises majority classes (green and orange), but neglects minority ones (blue and red). However, in Figure \ref{fig:case1}b, the combination of class-specific and generic transformers effectively distinguishes all four classes.

\subsection{A Case Study of BasicMotions}

\begin{figure}[t]
\begin{center}
\includegraphics[width=\linewidth]{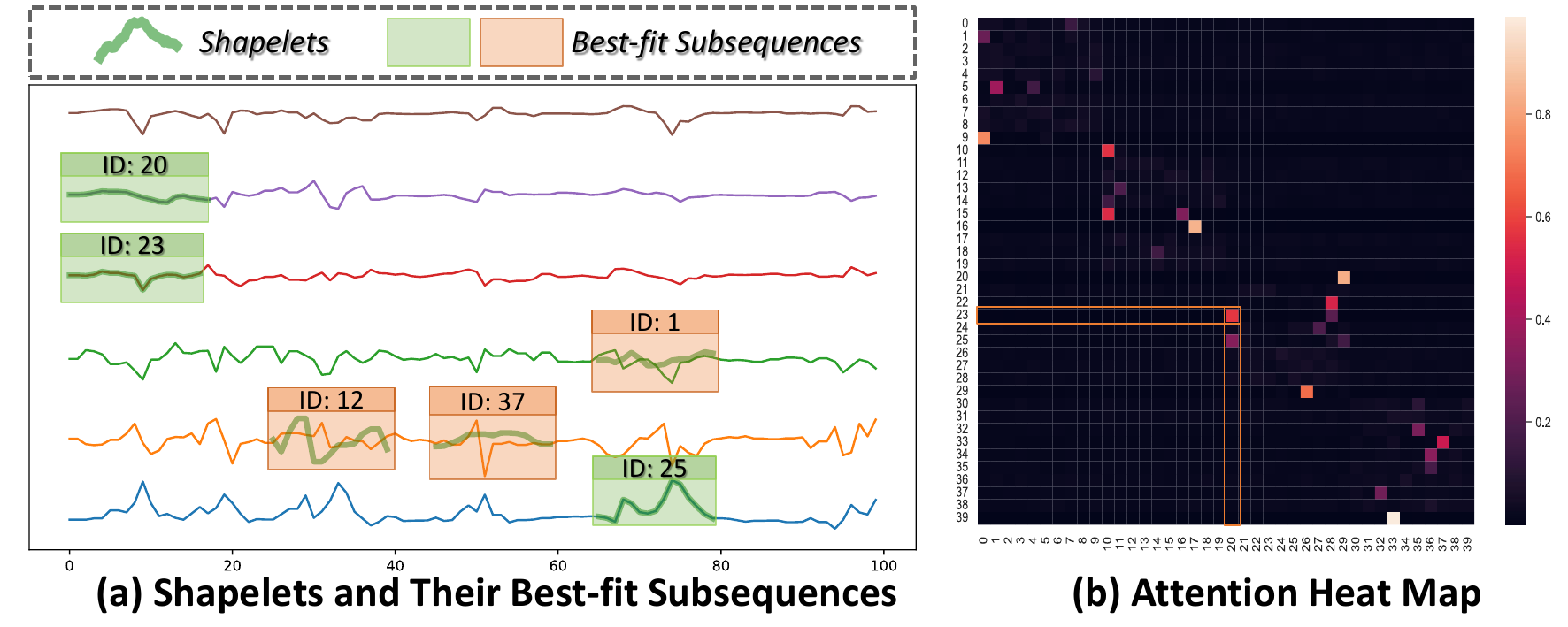}
\end{center}
\caption{(a) The green box depicts the top three shapelets, and the orange box displays three random shapelets from other classes, extracted in one random input time series of the 'walking' class in the BasicMotions dataset. (b) The attention heat map for all shapelets.}
\label{fig:case2}
\end{figure}

To interpret ShapeFormer results, we use the \textit{BasicMotions} dataset from the UEA archive \cite{uea}, focusing on human activity recognition with 4 classes (playing badminton, standing, walking, and running). Each class is associated with 6 variables, and 10 shapelets are set for analysis. Randomly selecting a 'walking' instance from the training set. Figure \ref{fig:case2}a showcases the top three shapelets for this class and three from others, highlighting ShapeFormer's ability to identify crucial subsequences across diverse locations and variables in the time series. Moreover, shapelets within the same 'walking' class tend to share greater similarity with best-fit subsequences than those from other classes.

In Figure \ref{fig:case2}b, the attention heat map for all 40 shapelets across 4 classes reveals that shapelets within the same class generally attain higher attention scores. For instance, $S_{20}$ and $S_{23}$ belonging to the 'walking' class show a small difference feature (Eq. \ref{eq:u}), resulting in higher attention scores. This enhanced attention allows the model to focus more on the correlation between shapelets within the same classes, thereby improving overall performance.

\section{Conclusion}
In this paper, we propose a novel Shapelet Transformer (ShapeFormer) for multivariate time series classification. It consists of dual transformer modules aimed at identifying class-specific and generic features within time series data. In particular, the first module discovers class-specific features by utilising discriminative subsequences (shapelets) extracted from the entire dataset. Meanwhile, the second transformer module employs convolution filters to extract generic features across all classes. The experimental results show that by combining both modules, our ShapeFormer has achieved the highest rank in terms of classification accuracy when compared to the SOTA methods. In future work, we intend to utilise the power of shapelets in many different time series analysis tasks such as forecasting or anomaly detection.

\bibliographystyle{ACM-Reference-Format}
\bibliography{shapeformer}

\appendix




\section{Baselines} 
\label{ap:baseline}
12 baseline methods are utilised in the comparative experiments, comprising two distance-based methods, a pattern-based algorithm, a feature-based algorithm, an ensemble method, three deep learning models, an attention-based model, and three transformer-based models. They all attained the SOTA performance described in the most recent research.

\begin{itemize}[noitemsep, nolistsep, leftmargin=*]
    \item \textit{EDI} and \textit{$DTW_D$} \cite{uea}: The two benchmark classifiers based on Euclidean Distance and dimension-dependent dynamic time warping.

    \item \textit{WEASEL+MUSE} \cite{weasel}: A classifier based on a bag-of-pattern approach demonstrated SOTA performance when compared with similar competitors for MTSC. We choose this algorithm as the representative baseline among pattern-based methods.

    \item \textit{MiniRocket} \cite{minirocket}: A feature-based method utilizes random convolutional kernels to discover features. It performs well in both univariate and multivariate time series classification.

    \item \textit{LCEM} \cite{lcem}: A hybrid ensemble method that integrates boosting, bagging, divide-and-conquer, and decision tree components. LCEM demonstrated superior performance when compared to other random forest methods. We choose it as a representative illustration of ensemble methods.

    \item \textit{MLSTM-FCNs} \cite{mlstm-fcns}: A deep learning method for MTSC which utilizes LSTM layer and stacked CNN layers to discover features.
 
    \item \textit{Tapnet} \cite{tapnet}:  A classifier constructs an attentional prototype network. Tapnet incorporates LSTM, and CNN to learn multi dimensional interaction features. We opt for it as another representative of the deep learning method.

    \item  \textit{Shapenet} \cite{shapenet}: Shapenet aims to learn representations of different shapelet candidates in a unified space and selects final shapelets by training a dilated causal CNN module followed by standard classification. This model can capture dependencies among variables.  We choose it as a representative of the shapelet-based method.

    \item  \textit{WHEN} \cite{when}: An attention-based method that learns heterogeneity by utilising a hybrid attention network, incorporating both DTW attention and wavelet attention. It achieved SOTA performance for MTSC on the UEA datasets.

    \item \textit{TST} \cite{TST}: A transformer-based framework for MTS representation learning. TST is considered as baseline method that takes the values at each timestamp as the input for the Transformer model. It gains great performance for many sequential tasks, such as regression, and classification.

    \item  \textit{ConvTran} \cite{ConvTran}: A transformer-based method for MTSC that proposed to improve the position embedding in the Transformer model by leveraging both absolute and relative position encoding.

    \item \textit{SVP-T} \cite{svp-t}: A method uses clustering to identify time series subsequences and employs them as inputs for the Transformer, enabling the capture of long- and short-term dependencies among subseries. It achieved SOTA performance for MTSC. We choose it as another representative of the transformer-based method.
\end{itemize}

\section{Selected Hyperparameters }
\label{sec:ab-para}
In this section, we follow the setting mentioned in Section \ref{sec:hyper} to tune the hyperparameter of window size and number of shapelets per class. In Table \ref{tab:parameter}, we provide the selected window size and number of shapelets for each class on 30 UEA MTSC datasets. 

\begin{table}[t]
\caption{The selected window size and number of shapelets for each UEA MTSC dataset.}
\begin{adjustbox}{width=\linewidth}
\begin{tabular}{@{}lcc@{}}
\toprule
Dataset                   & Window size & Number of shapelets (per class) \\ \midrule
ArticularyWordRecognition & 50          & 10                   \\
AtrialFibrillation        & 100         & 3                    \\
BasicMotions              & 100         & 10                   \\
CharacterTrajectories     & 50          & 3                    \\
Cricket                   & 200         & 30                   \\
DuckDuckGeese             & 10          & 100                  \\
ERing                     & 50          & 100                  \\
EigenWorms                & 10          & 10                   \\
Epilepsy                  & 20          & 30                   \\
EthanolConcentration      & 200         & 100                  \\
FaceDetection             & 10          & 10                   \\
FingerMovements           & 20          & 30                   \\
HandMovementDirection     & 200         & 100                  \\
Handwriting               & 20          & 30                   \\
Heartbeat                 & 200         & 100                  \\
InsectWingbeat            & 10          & 30                   \\
JapaneseVowels            & 10          & 1                    \\
LSST                      & 20          & 10                   \\
Libras                    & 10          & 30                   \\
MotorImagery              & 100         & 30                   \\
NATOPS                    & 20          & 1                    \\
PEMS-SF                   & 50          & 10                   \\
PenDigits                 & 4           & 10                   \\
PhonemeSpectra            & 20          & 30                   \\
RacketSports              & 10          & 10                   \\
SelfRegulationSCP1        & 100         & 100                  \\
SelfRegulationSCP2        & 100         & 100                  \\
SpokenArabicDigits        & 10          & 100                  \\
StandWalkJump             & 10          & 100                  \\
UWaveGestureLibrary       & 10          & 10                   \\ \bottomrule
\end{tabular}
\end{adjustbox}
\label{tab:parameter}
\end{table}

\section{Comparison with Time Series Representation Methods}
We conducted an experiment to compare the average accuracy of our ShapeFormer against SOTA representation methods, specifically TST \cite{TST}, PatchTST \cite{patchTST}, TimesNet \cite{timesnet}, and GPT2 \cite{gpt2} (as shown in Table \ref{tab:compare_SOTA_rl}). By adhering to the GPT2 protocol across 10 datasets, our method outperforms the others on all datasets, achieving an average accuracy of 0.773.

\begin{table}[H]
\caption{Comparison between our proposed ShapeFormer and SOTA time series representation methods.}
\begin{adjustbox}{width=\linewidth}
\begin{tabular}{|l|l|l|l|l|l|}
\hline
                   & TST  & GPT2   & TimesNet  & PatchTST  &  Ours \\ \hline
Averaging Accuracy & 0.736 & 0.740 & 0.736    & 0.679    & \textbf{0.773}     \\ \hline
\end{tabular}
\end{adjustbox}
    \label{tab:compare_SOTA_rl}
\end{table}

\end{document}